%% file: main.tex
\documentclass{article}

\usepackage[preprint]{neurips_data_2021}
\usepackage[utf8]{inputenc} % allow utf-8 input
\usepackage[T1]{fontenc}    % use 8-bit T1 fonts
\usepackage{hyperref}       % hyperlinks
\usepackage{url}            % simple URL typesetting
\usepackage{booktabs}       % professional-quality tables
\usepackage{amsfonts}       % blackboard math symbols
\usepackage{nicefrac}       % compact symbols for 1/2, etc.
\usepackage{microtype}      % microtypography
\usepackage{xcolor}         % colors
\usepackage[textsize=footnotesize]{todonotes}
\usepackage{subcaption}
\usepackage{enumitem}

\usepackage{setspace}
\usepackage{multicol}

\setcitestyle{square,numbers} % To use numbers in citation

\newcommand{\TaBERT}{{\sc TaBert}}
\newcommand{\TAPAS}{{\sc TaPas}}
\newcommand{\RCI}{{\sc RCI}}

\usepackage{amssymb}% http://ctan.org/pkg/amssymb
\usepackage{pifont}% http://ctan.org/pkg/pifont
\newcommand{\cmark}{\ding{51}}%
\newcommand{\xmark}{\ding{55}}%

\title{\sys: Question Answering Dataset over Complex Tables in the Airline Industry}
\input{macros}

\author{
Yannis Katsis$^{1}$,
Saneem Chemmengath$^{1}$,
Vishwajeet Kumar$^{1}$,\\
\textbf{Samarth Bharadwaj$^{1}$,
Mustafa Canim$^{1}$,
Michael Glass$^{1}$,
Alfio Gliozzo$^{1}$,} \\
\textbf{Feifei Pan$^{2}$,
Jaydeep Sen$^{1}$, 
Karthik Sankaranarayanan$^{1}$, 
Soumen Chakrabarti$^{3}$} \\
$^{1}$IBM Research  
$^{2}$Rensselaer Polytechnic Institute
$^{3}$IIT Bombay\\
yannis.katsis@ibm.com,
saneem.cg@in.ibm.com, 
vishk024@in.ibm.com,\\
samarth.b@in.ibm.com,
mustafa@us.ibm.com,
mrglass@us.ibm.com,  
gliozzo@us.ibm.com, \\   
panf2@rpi.edu,
jaydesen@in.ibm.com, 
kartsank@in.ibm.com, 
soumen.chakrabarti@gmail.com\\
    }

\begin{document}
\setlist[itemize]{leftmargin=15pt}
\setlist[enumerate]{leftmargin=15pt}
\maketitle

\begin{abstract}
\input{0_abstract-new}
\end{abstract}

\input{1_introduction}
\input{2_related}

\input{3_dataset}

\input{4_experiments}
\input{5_conclusion}

% 100 test
% 100 dev
% 315 -> 10% 30% 50% 70% 

% Intro, abstract, conc -> Mustafa, Michael, Yannis
% Related work -> Feifei
% Dataset -> Yannis
% Experiments -> IRL
% Huge Analysis section -> 

\bibliographystyle{acl_natbib}
\bibliography{anthology,custom}

% \section{TODOs (To be removed):}
% \begin{enumerate}
%   \item
%     Dataset documentation and intended uses. 
    
%   \item 
%   Recommended documentation frameworks include datasheets for datasets, dataset nutrition labels, data statements for NLP, and accountability frameworks. 
%   \answerTODO{One of these }
%   \item 
%   URL to website/platform where the dataset/benchmark can be viewed and downloaded by the reviewers. \todo{done.}
    
%   \item
% a persistent dereferenceable identifier (e.g. a DOI minted by a data repository or a prefix on identifiers.org) for datasets, or a code repository
% \end{enumerate}

%%%%%%%%%%%%%%%%%%%%%%%%%%%%%%%%%%%%%%%%%%%%%%%%%%%%%%%%%%%%
\newpage

\appendix

{\onecolumn \centering \Large \bfseries \sys: Question Answering Dataset over Complex Tables in the Airline Industry \\ (Appendix) \\
 \par \medskip}

\section{Annotation Tool}
\begin{figure}[htb!]
    \centering
    \includegraphics[width=\textwidth]{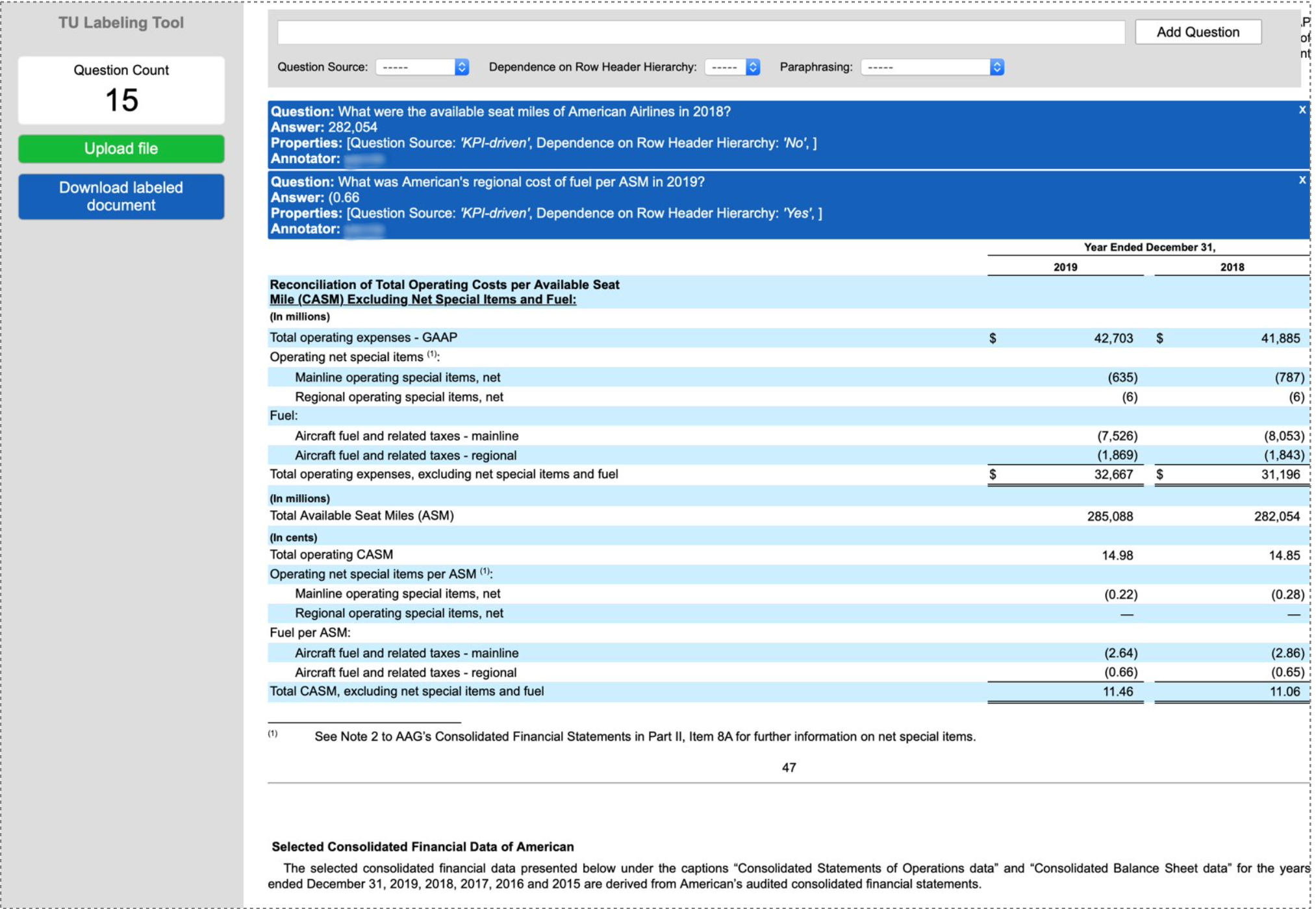}
    \caption{Annotating questions for a table using custom annotation tool}
    \label{fig:annotation_tool}
\end{figure}

In figure \ref{fig:annotation_tool}, we present annotation page of our custom table question answering annotation tool. Annotators were asked to generate question for the given table and annotate various properties of the question like whether the question is KPI-driven or table-driven etc.

\end{document}

%% file: macros.tex
\newcommand{\dataset}{AIT-QA} 
\newcommand{\sys}{AIT-QA}

%% file: 0_abstract-new.tex
Recent advances in transformers have enabled Table Question Answering (Table QA) systems to achieve high accuracy and SOTA results on open domain datasets like WikiTableQuestions and WikiSQL. Such transformers are frequently pre-trained on open-domain content such as Wikipedia, where they effectively encode questions and corresponding tables from Wikipedia as seen in Table QA dataset. However, web tables in Wikipedia are notably \emph{flat} in their layout, with the first row as the sole column header. The layout lends to a relational view of tables where each row is a tuple. Whereas, tables in domain-specific business or scientific documents often have a much more \emph{complex} layout, including hierarchical row and column headers, in addition to having specialized vocabulary terms from that domain. 

To address this problem, we introduce the domain-specific Table QA dataset \sys{} (Airline Industry Table QA). The dataset consists of 515 questions authored by human annotators on 116 tables extracted from public U.S. SEC filings\footnote{SEC Filings publicly available at: https://www.sec.gov/edgar.shtml} of major airline companies for the fiscal years 2017-2019. We also provide annotations pertaining to the nature of questions, marking those that require hierarchical headers, domain-specific terminology, and paraphrased forms.
Our zero-shot baseline evaluation of three transformer-based SOTA Table QA methods - TaPAS (end-to-end), TaBERT (semantic parsing-based), and RCI (row-column encoding-based) - clearly exposes the limitation of these methods in this practical setting, with the best accuracy at just 51.8\% (RCI). We also present pragmatic table pre-processing steps used to pivot and project these complex tables into a layout suitable for the SOTA Table QA models.

%% file: 1_introduction.tex
\section{Introduction}

%or Google NQ dataset~\cite{sigir20}. %Yannis: I commented out GoogleNQ as according to our related work section, it as a table retrieval dataset. 
The tabular data format is commonly used in digital documents such as PDFs and HTMLs to store semi-structured information~\cite{DBLP:canim2019, ZhangAdHoc, pasupat-liang-2015-compositional}. Due to the rich content found in tables, many studies have been carried out on extracting information out of the tables~\cite{BurdickDEKW20} and leveraging it for various NLP tasks, such as answering questions over tables~\cite{cafarella2009data, yan2017contentbased, shraga2020ad, sigir20}. The quality of answers depends on first, high quality extraction of tables out of documents (aka \emph{table extraction}); second, retrieval of relevant tables for a given natural language question or query keyword (aka \emph{table retrieval}); and finally, identification of the relevant cells over the retrieved tables (aka \emph{table QA}). Most recently, transformer-based pre-trained architectures such as \TaBERT{} \cite{tabert}, \TAPAS{} \cite{tapas}, and \RCI{} \cite{rci} have been proposed to tackle the table QA task by identifying table cells containing the answer to a given question. These advanced models have been shown to perform well in answering questions over tables. However, most of these studies claim high accuracy in Table QA by evaluating the proposed techniques on open in-domain datasets, such as WikiTableQuestions~\cite{pasupat2015compositional} and WikiSQL dataset~\cite{zhongSeq2SQL2017}; both built on top of Wikipedia tables.

%Our analyses show that open domain table QA datasets are far simpler than tables that are used in domain specific documents such as finance, in terms of both table structure as well as the technical vocabulary used in these documents. Two sample tables are given in Figure~\ref{fig:example} where the first one is extracted from a domain specific document whereas the second one shows a table from WikiTableQuestions dataset which is an open domain dataset for Table QA task. Nested structure of row headers as well as the hierarchical column headers makes it harder to find the answer of a given question even for human experts. Our experiments (reported in Section~\ref{sec:experiments}) show that even the most advanced transformer-based pre-trained models struggle to comprehend the layout of these domain specific tables and find the right answer of natural language questions. Clearly, the lack of a domain specific dataset and benchmark for the table QA task plays a major role in this incompetency as these models are all evaluated by open domain benchmarks where tables contain limited vocabulary terms and simple column/row layout.

\begin{figure}[t!]
\centering
\includegraphics[width=.8\textwidth]{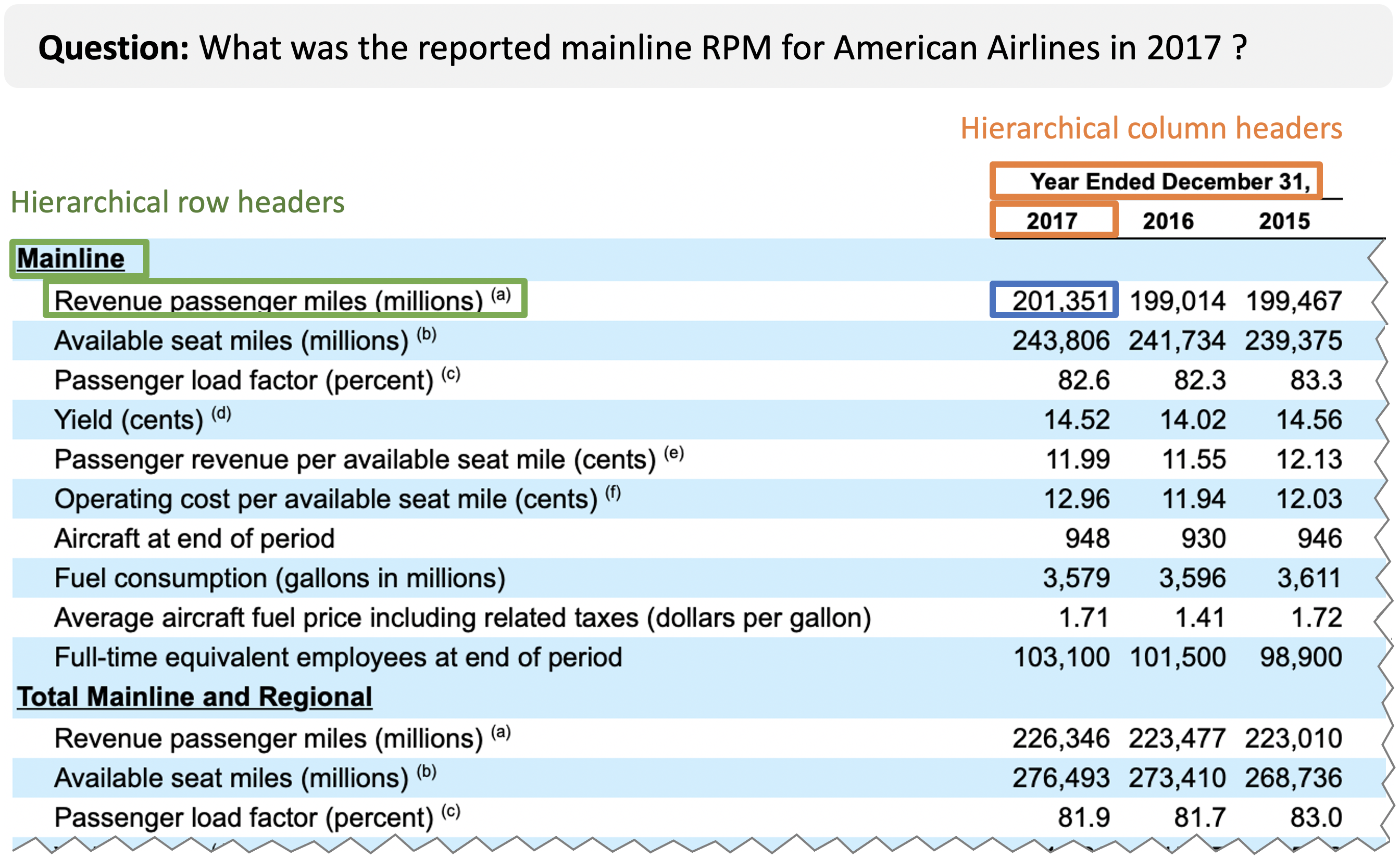}
%{images/complex_table.png}
\caption{A question-table pair in \sys, showcasing the complex structure of tables in the dataset. The table cell containing the answer is highlighted in blue and its hierarchical column and row headers are highlighted in orange and green, accordingly.}
\label{fig:example}
%This figure is available in ppt in the box folder. Please use it to edit and make this better.
\end{figure}

However, based on our prior experience in table processing \cite{BurdickDEKW20}, open domain web tables typically exhibit a much simpler structure than tables found in scientific or business documents. For instance, consider the sample question-table pair from our proposed airline industry dataset shown in Figure~\ref{fig:example}. The table contains both column headers and row headers (i.e., descriptors of the contents of columns and rows, respectively) and both of them are hierarchical in nature. Moreover, answering the question often requires reasoning on these complex column and row header hierarchies. For instance, finding the requested mainline Revenue Passenger Miles (which are contained in the cell highlighted in blue) requires understanding and leveraging the fact that the cell has two hierarchical row headers "Mainline" and "Revenue passenger miles" (shown in green). Ignoring the row headers or not reasoning on the entire row header hierarchy may lead to the wrong result. For instance, if we simply searched for cells with a flat row header containing "Revenue Passenger Miles", we may mistakenly return the value 226,346 appearing further down the table. This value indeed corresponds to Revenue Passenger Miles (RPMs) but these are the RPMs of the entire \emph{mainline and regional operations} of the airline, instead of only the \emph{mainline} operations requested by the question. In contrast, web tables appearing in open domain Table QA datasets, such as WikiTableQuestions or WikiSQL, exhibit significantly simpler structures. Such tables do not contain any row headers at all and only have a single column header, closely resembling relational database tables.

Moreover, while open domain datasets capture common entities, such as locations, person names, etc, which often appear in Wikipedia articles, they often lack domain-specific vocabulary that one encounters in scientific or business documents (such as the ``Revenue Passenger Miles" above).  

Our experiments (reported in Section~\ref{sec:experiments}) show that even the most advanced transformer-based pre-trained models struggle to understand the layout of these domain specific tables and find the right answer of natural language questions. We argue that the lack of a domain-specific datasets for Table QA is partially responsible for this incompetency, as these models are all evaluated on open domain datasets, where tables contain limited specialized vocabulary and adopt a simple column/row layout.

To address this gap in the Table QA literature and assist the community in improving the performance of Table QA approaches for domain specific use cases, in this paper we propose a domain specific dataset, where tables are extracted from financial documents in the airline industry. The majority of the tables exhibit a complex structure, including hierarchical row and column headers, large amounts of numerical values, as well as technical vocabulary terms specific to the airline industry. To the best of our knowledge, this is the first dataset tailored for the Table QA task that includes and explicitly encodes such complex table layouts, domain-specific table contents, as well as questions manually created by human annotators to test Table QA algorithms.

This work makes the following contributions:
\begin{itemize}
  \item \textbf{A complex and domain specific Table QA dataset} called
  \emph{\sys{} (Airline Industry Table QA)}, created by human annotators based on 10-K financial reports of major airline companies. The questions are created based on the content of tables appearing in the 10-K reports, as well as KPIs (Key Performance Indicators); i.e., important metrics commonly used by analysts in the airline industry.
  \item \textbf{Experimental evaluation of state-of-the-art Table QA models} applied on \sys{}, demonstrating that high performance of open domain datasets with simple table structures does not guarantee similar performance on domain-specific datasets containing complex tables, further motivating the need for a domain-specific TableQA dataset.
  \item \textbf{A novel data pre-processing technique for existing Table QA models}, which improves their performance on datasets with complex table structures. This is accomplished by translating complex table structures (including hierarchical row and column headers) to simpler table structures resembling the structure of the tables on which such approaches have been trained. 
\end{itemize}

The paper is structured as follows: We start by reviewing existing datasets on answering questions over tables in Section~\ref{sec:related}. We then describe our proposed \sys{} dataset and how it was constructed in Section~\ref{sec:approach}. Section~\ref{sec:experiments} describes experimental results of state-of-the-art models over \sys{} and finally Section~\ref{sec:conclusion} concludes the paper.\footnote{All resources are available at \url{https://github.com/IBM/AITQA}.} 
%Datasheet and Neurips checklist are available as supplementary material.}
\label{sec:intro}

% Why Table QA is important
% There are open datasets but they are not domain specific and they contain tables with simple structure
% How domain specific Table QA datasets are different from open datasets
% How table structures differ
% What we propose in this paper which is different from existing datasets
% Show example of question-table pair
% Explain that the goal is to use the dataset for out-of-domain testing

%% file: 2_related.tex
\section{Related Work}
\label{sec:related}
% \subsection{QA over text benchmarks}
% Traditional QA systems were mainly trained on large-scale open domain text corpora.
Existing work on leveraging tables to answer questions has in general focused on two distinct tasks: (a)~\emph{Table retrieval}; i.e., given a corpus of tables, identifying the table that contain the answer to a question, and (b)~\emph{Table QA}; i.e., given a single table containing the answer to the question, finding the answer of that question. While our dataset is tailored for the Table QA task, we next summarize existing datasets proposed for either of the two aforementioned tasks. 

%Most existing Table QA benchmarks focus on either answering natural language questions on a single table or retrieving multiple tables using keyword-based queries. These table QA benchmarks are mainly open domain datasets based on the Wikipedia corpus, such as WebTables \cite{webtables}, WikiTables~\cite{wikitable}, WikiTableQuestions~\cite{wikitablequestions}, WikiSQL~\cite{wikisql}, etc. In the following sections, we discuss some popular benchmarks for table retrieval and QA over tables in details. 

\subsection{Table Retrieval Datasets} 
%todo webtables, 
WebTables ~\cite{webtables} is one of the largest table corpora that is publicly available for table retrieval, with approximately 14.1 billion HTML tables crawled from the English text documents in the main index of Google. Most state-of-the-art models use keyword based queries \citep{zhang2018ad, Pimplikar_2012, gao2017scientific}.
For example, \citet{zhang2018ad} evaluated their proposed model using \emph{WikiTables} \citep{wikitable} as table corpus. WikiTables is a table dataset with about 1.6 million relational tables generated from 154 million Web tables. \citet{zhang2018ad} also released a hand-crafted query dataset with 60 keyword queries, such as `phases of the moon' and `science discoveries', together with 3,120 annotated tables from WikiTables. This dataset was later used by many other researchers, including the current state-of-the-art model for table retrieval \citep{sigir20}, published in 2020. 

\citet{sigir20} process the \emph{Natural Questions (NQ)} corpus \citep{47761} and release a new table retrieval dataset, the \emph{GNQtables} dataset, which contains 789 natural language questions over 74,224 tables. The NQ corpus is designed for general QA tasks, with more than 323,000 examples of real queries from the Google search engine. The answers to the questions in NQ are generated from Wikipedia articles. Except from the GNQtables dataset, a new derivative dataset extending NQ dataset has been contributed by \citet{herzig2021open}, which identifies 12,000 natural language questions with answers in tables. 

Other published datasets for table retrieval include WebQueryTables \citep{yan2017contentbased}, TableArXiv \citep{gao2017scientific}, Web Data Commons (WDC) table corpus \citep{webdatacommons2016}, etc.

\subsection{Table QA Datasets}

In Table QA studies (which are the most relevant to our work), the most commonly used datasets include WikiTableQuestions, WikiSQL, and TabMCQ. \emph{WikiTableQuestions} \citep{wikitablequestions} is a dataset containing 22,033 question-table pairs over 2,108 Wikipedia tables.
%Each table in WikiTableQuestions contains at least five columns and eight rows. The questions are human-curated natural language questions prepared by two Amazon Mechanical Turks.
\emph{WikiSQL} \citep{wikisql} is also a Wikipedia-based dataset with 80,654 hand-annotated natural language questions over a corpus of 24,241 Wikipedia tables. On the other hand, \emph{TabMCQ} \citep{jauhar2016tabmcq} does not build on Wikipedia, but instead contains 500 multiple choice questions over 70 manually-curated general knowledge tables created from the Regents 4th-grade exam. While this dataset is domain-specific, the included tables have a very peculiar structure, with table rows containing entire natural language sentences that have been appropriately split into columns. In our experience, the resulting tables capture a very special case and are not representative of tables appearing in most domains. More recently, \citep{nan2021fetaqa} proposed FeTaQA, a new free-form Table QA dataset. Compared to prior datasets, the main novelty of FeTaQA lies in the structure of the answers, which are long free-form sentences (in contrast to the very short answers found in prior datasets). FeTaQA is also build on Wikipedia, containing 10,330 unique Wikipedia-based instances covering 16 different topics.

%The data of FeTaQA are stored as records of the form \{\emph{table, question, free-form answer, supporting  table cells}\}.

Finally, the last couple of years saw the introduction of three multi-hop QA datasets: \emph{HybridQA} \citep{chen-etal-2020-hybridqa}, \emph{OTT-QA} \citep{chen2021ottqa}, and \emph{TAT-QA}~\citep{zhu2021tatqa}. These datasets are designed to accommodate a special scenario, where finding answers to natural language questions requires reasoning not only on tables but across both tables and associated text. Out of them, HybridQA and OTT-QA are both based on Wikipedia. On the other hand, TAT-QA, is based on data extracted from financial reports, making it the most similar dataset to our proposed \sys{}.

However, while the TAT-QA paper mentions the existence of complex table structures (including row headers) in financial tables, the resulting dataset does not include explicit annotations of row and column headers (not to mention hierarchies thereof). Without explicit annotations of such headers, not only is it hard to understand the complexity of the included tables (for instance the results of a manual table inspection included in the Appendix of \citep{zhu2021tatqa} points to the absence of column header hierarchies in TAT-QA), but it also makes it harder to understand the effect of table structure complexity on the performance of TableQA algorithms. Instead, our proposed \sys{} treats hierarchical column and row headers as first-class citizens and is to the best of our knowledge the first Table QA dataset that contains explicit annotations of such complex table structures. This provides three advantages: First, it is the first dataset with guaranteed coverage of such complex structures. Second, it enables a principled analysis of the effect of complex table structures on Table QA algorithms. Finally, it opens up the path of a principled separation of the column/header identification task from the table QA task, thus leveraging prior work on table header identification \cite{nishida, jing2012, xilun-rectanglemining, koci2016cell}.  

Table \ref{tab:dataset-comparison} summarizes the aforementioned Table QA datasets. As discussed above, apart from our work, all existing TableQA datasets with the exception of TabMCQ and TAT-QA are based on Wikipedia tables. Moreover, only our work contains and explicitly encodes hierarchical row and column headers. While TAT-QA may contain a subset of such complex structures, they are not encoded as such. Moreover, all other datasets do not consider row headers at all and consider only flat column headers (i.e., cases where the single top row of the table contains column headers).

% -open domain benchmarks: wikisql wikitablequestions  wikitables google
% -benchmarks designed for a specific task: tabmcq MQA-QG OTT-QA FeTaQA
%- NL2SQL datasets: https://towardsdatascience.com/ai-enabled-conversations-with-analytics-tables-66a10c9a3d05
% \begin{table*}
% \centering
% \begin{tabular}{@{}lllccc@{}}
% \toprule
% \textbf{Dataset} & \textbf{Year} & \textbf{Hops} & \textbf{Domain} & \textbf{Column} & \textbf{Row}\\
% & & & & \textbf{Headers} & \textbf{Headers}\\
% \midrule
% WikiTableQuestions \cite{wikitablequestions} & 2015 & Single-hop & Wikipedia & Flat & None\\
% TabMCQ \cite{jauhar2016tabmcq} & 2016 & Single-hop & ? & ? & None\\
% WikiSQL \cite{wikisql} & 2017 & Single-hop & Wikipedia & Flat & None\\
% FeTaQA \cite{nan2021fetaqa} & 2021 & Single-hop & Wikipedia & Flat & None\\
% \midrule
% HybridQA \cite{chen-etal-2020-hybridqa} & 2020 & Multi-hop & Wikipedia & Flat & None\\
% OTT-QA \cite{chen2021ottqa} & 2021 & Multi-hop & Wikipedia & Flat & None\\
% TAT-QA \cite{zhu2021tatqa} & 2021 & Multi-hop & Finance & Flat & ?\\
% \midrule
% \dataset{} (ours) & 2021 & Single-hop & Airlines & Hierarchical & Hierarchical \\
% \bottomrule
% \end{tabular}
% \caption{\label{tab:dataset-comparison} Comparison of Table QA Datasets}
% \end{table*}

\begin{table*}
\centering
\begin{tabular}{@{}llcccc@{}}
\toprule
\textbf{Dataset} & \textbf{Year} & \textbf{Table only} & \textbf{Wikipedia} & \textbf{Hierarchical} & \textbf{Hierarchical} \\
 &  &  &  & \textbf{Column} & \textbf{Row} \\
 &  &  &  & \textbf{Headers} & \textbf{Headers} \\ \midrule
WikiTableQuestions \cite{wikitablequestions} & 2015 & \cmark & \cmark & \xmark & \xmark \\
TabMCQ \cite{jauhar2016tabmcq} & 2016 & \cmark & \xmark\hspace*{0.1cm}(Science Exam) &  \xmark & \xmark\\
WikiSQL \cite{wikisql} & 2017 & \cmark & \cmark & \xmark & \xmark \\
FeTaQA \cite{nan2021fetaqa} & 2021 & \cmark & \cmark & \xmark & \xmark \\ \midrule
HybridQA \cite{chen-etal-2020-hybridqa} & 2020 & \xmark & \cmark & \xmark & \xmark \\
OTT-QA \cite{chen2021ottqa} & 2021 & \xmark & \cmark & \xmark & \xmark \\
TAT-QA \cite{zhu2021tatqa} & 2021 & \xmark & \xmark\hspace*{0.1cm}(Finance) & \xmark & \xmark \\ \midrule
\dataset{} (this work) & 2021 & \cmark & \xmark\hspace*{0.1cm}(Airlines) & \cmark & \cmark \\ \bottomrule
\end{tabular}
\caption{\label{tab:dataset-comparison} Comparison of \sys{} to other Table QA datasets}
\end{table*}

%% file: 3_dataset.tex
\defcitealias{edgar}{U.S. Securities and Exchange Commission}
\defcitealias{sp500}{Wikipedia}

\section{Dataset}
\label{sec:approach}

We next explain the process we followed to generate our dataset, starting from data acquisition and data preparation and continuing to question annotation and table header identification.

\paragraph{Data Acquisition.} \dataset{} is based on 10-K forms; comprehensive annual reports that publicly traded companies file with the U.S. Securities and Exchange Commission (SEC). For this dataset, we focused on the airline industry and retrieved recent 10-K forms of all 5 airlines included in the Standard \& Poor's 500 (S\&P 500) stock market index (\citetalias{sp500}). The covered airlines include (stock ticker symbols shown in parenthesis): Alaska Air Group (ALK), American Airlines Group (AAL), Delta Air Lines Inc. (DAL), Southwest Airlines (LUV), and United Airlines Holdings (UAL). The 10-K forms were downloaded through the SEC EDGAR online system (\citetalias{edgar}) in HTML form. The dataset files are available at \url{https://github.com/IBM/AITQA}. 

\paragraph{Data Preparation and Cleaning.} While the downloaded 10-K forms encode tables using standard HTML tags, the tabular structures are designed with human consumption in mind. As such, table rows/columns/cells are used to allow for the table to be neatly rendered on the screen and/or paper and they do not always correspond to the table's logical structure. In particular, we found that tables in the downloaded 10-K forms contain extraneous rows/columns (introduced to allow for more space between table elements). Moreover, the contents of a single logical cell are often split into multiple physical cells, to allow for better vertical alignment of the information within a table. For instance, cells containing a currency symbol and negative monetary amounts such as $\$(1,234)$, are often split into three physical cells
\begin{tabular}{@{}|l|l|l|@{}}
\hline
$\$$ & $(1,234$ & $)$\\
\hline
\end{tabular}
so that the currency symbols and numbers align with other similar contents across table rows. To separate these formatting decisions from the logical structure of the table, we post-processed the downloaded HTML files to remove extraneous rows and columns and merge back together components of logical cells that were split into multiple cells. Processing was done through a combination of scripts and manual error correction. 

\paragraph{Question Annotation.} The cleaned 10-K forms were given to 8 co-authors of this paper to generate question-answer pairs over tables appearing on the forms. To capture questions of particular interest to domain experts in the domain, while ensuring a diversity of question topics, we asked annotators to provide two types of questions:
\begin{itemize}
    \item \emph{KPI-driven questions:} These are questions that inquiry about Key Performance Indicators (KPIs), which are metrics of particular interest to analysts in the airline industry. While creating these questions, annotators were provided with a list of KPIs along with common synonyms to ensure that the questions capture not only the topic of interest but also use the respective vocabulary. To generate these questions, annotators were instructed to search the document for mentions of KPIs appearing within tables and create corresponding questions. Thirteen KPIs were used in total, with each of them having three variants, depending on whether it referred to the airlines' mainline operations, its regional operations, or the combination thereof.
    \item \emph{Table-driven questions:} While KPI-driven questions capture the common metrics inquired by analysts, they can be limiting for two main reasons: First, there is a limited number of KPIs and second, given their importance in the domain, they often appear within a small set of tables. As a result, limiting ourselves only to such questions would lead to a non-diverse dataset. To avoid this issue, annotators were asked to also provide questions that inquired about other concepts appearing within the input tables. To create such questions, annotators had to browse through the tables in the input documents and write questions that could be answered by them.
\end{itemize}
%\todo{what measures, if any, were taken to ensure that there are enough queries that need to navigate complex row/column headers?}
After an initial set of question-answer pairs was collected, annotators were also asked to generate paraphrases. While creating the paraphrased questions, annotators were given access to the set of question-answer pairs collected in earlier stages and asked to pick a subset of questions to paraphrase. This leads to the second major dimension along which questions in our dataset can be classified:
\begin{itemize}
    \item \emph{Original questions:} Questions collected during the initial stages of the annotation.
    \item \emph{Paraphrased questions:} Questions generated as paraphases of original questions.
\end{itemize}
Finally, in all stages of the annotation process, annotators were also asked to keep track of additional metadata indicating whether a question relied on the hierarchy of row headers to be answered. A question relies on the hierarchy of row headers when in order to be unambiguously answered, one has to not only see the row header that appears on the same row as the answer, but also on the higher levels of the hierarchy. For instance, the question in Figure~\ref{fig:example} depends on the row header hierarchy, as ignoring the hierarchy may lead to an incorrect answer, as explained in the introduction. Based on these metadata, questions in the dataset can be differentiated across a third dimension into:
\begin{itemize}
    \item \emph{Row header hierarchy questions:} Questions whose answer relies on the row header hierarchy.
    \item \emph{No row header hierarchy questions:} Questions whose answer does not rely on the row header hierarchy.
\end{itemize}

For each question-answer pair, annotators provided the question, the table cell where the answer appears, as well as metadata indicating the classification of the question along the three aforementioned dimensions. For the first version of the dataset, we focus on \emph{lookup} questions - i.e., questions where the answer appears within table cells and does not require aggregate operations (such as min/max/sum/count) to be returned \cite{rci}, leaving the expansion of the dataset with aggregate questions as future work. Annotation was carried out using a custom-built Table QA annotation tool (screenshot in Supplementary material). Finally, the collected question-answer pairs and associated metadata were subsequently reviewed by other annotators to verify their validity and correct minor issues, such as typos or associated metadata.

\paragraph{Hierarchical Column/Row Header Identification.} To identify column and row headers of tables, we leveraged Table Understanding technology incorporated in IBM Watson Discovery \cite{ibmwd}. Table Understanding allows among others identifying for each body (i.e., non-header cell), the set of column headers and row headers that describe the cell \cite{ibmwd-tu}. Table Understanding allows identifying both column and row header hierarchies, as described above. The identified header hierarchies are included as part of the dataset so that they can be leveraged by Table QA models.

\begin{table*}
\centering
\begin{subfigure}{0.7\textwidth}
\centering
\begin{tabular}{lc}
\toprule
\textbf{Question Type} & \textbf{Count (\%)}\\
\midrule
KPI-driven questions & 145 (28\%)\\
Table-driven questions & 370 (72\%)\\
\midrule
Original questions & 441 (86\%)\\
Paraphrased questions & \hspace*{0.1cm}76 (15\%)\\ 
\midrule
Row header hierarchy questions & 146 (28\%)\\
No row header hierarchy questions & 369 (72\%)\\
\bottomrule
\end{tabular}
\caption{Breakdown of questions across 3 dimensions}
\label{tab:stats-breakdown}
\end{subfigure}
\begin{subfigure}{0.28\textwidth}
\centering
\begin{tabular}{lc}
\toprule
\textbf{Property} & \textbf{Value}\\
\midrule
Documents & 13 \\
Companies & 5 \\
Date range & 2017-19\\
Questions & 515 \\
Tables & 116 \\
\bottomrule
\end{tabular}
\caption{Other dataset properties}
\label{tab:stats-other}
\end{subfigure}
\caption{Dataset Statistics}
\label{tab:stats}
\end{table*}

\paragraph{Dataset Statistics.} Statistics of the resulting dataset are shown in Table \ref{tab:stats}. Table \ref{tab:stats-breakdown} shows the breakdown of questions along the three aforementioned dimensions, while Table \ref{tab:stats-other} lists other properties of the dataset, including number of 10-K forms, companies, and tables.

%% file: 4_experiments.tex
\section{Experimental Evaluation}
\label{sec:experiments}
To analyze the effect of \sys's domain-specific complex tables to existing Table QA approaches, we next provide a comprehensive evaluation of state-of-the-art Table QA models when applied on it.

%In this section we provide a comprehensive evaluation and performance comparison of multiple state-of-the-art Table QA models on our newly introduced dataset \sys. We begin by describing the TableQA models and the experimental setup, followed by a detailed analysis of the performance of each baseline on \sys. 

\subsection{Baseline methods}
\label{baselines}
% Write about TaBERT, TAPAS, RCI models, how it was trained, codebase.
We evaluate three Table QA systems - \textbf{RCI}~\citep{rci}, \textbf{TaBERT}~\cite{tabert}, and \textbf{TaPaS}~\cite{tapas} - selected as being representative of most of the existing Table QA approaches. 

TaBERT is a table and question encoder specifically designed for Table QA. The encoder is used to predict a logical form in an encoder-decoder approach \citep{wikitable,mapo,nsm,nsm_type_constraint}. This logical form is then executed over the table and provides the final answer to the original question. TaBERT employs a BERT~\cite{bert} encoder and an LSTM decoder (which generates the logical forms~\cite{nsm}) and is trained using reinforcement learning~\cite{mapo}. A content snapshot heuristic is used to handle tables that are too large for the BERT encoder. 

On the other hand, both TaPaS~\cite{tapas} and RCI~\cite{rci} treat Table QA as classification problem and skip generating logical forms. In the case of TaPaS, during the pre-training phase, it jointly learns representations for natural language sentences and structured tables, which makes it is suitable for table question answering. Then, during the fine tuning phase, it follows a two step procedure, where it first selects relevant cell/cells and then optionally applies additional operators to the selected cells. RCI splits tables into rows and columns and carries out inference on them separately. To this end, it employs a row predictor identifying the row containing the answer and a column predictor identifying the corresponding column.
%It attempts to select the correct column, followed by correct row from the table given a natural language query, thus identifying the correct answer cell. 
Two separate BERT models are used for generating column/row representations and query representations. 
In all three systems, tables and questions are encoded using transformers \cite{transformer}.

\subsection{Experimental Setup}
We attempt to test whether high Table QA performance reported on open domain tables translates to \sys{} dataset. All three Table QA models are pre-trained on the larger WikiSQL \cite{wikisql} train split and tested on \sys{} without any hyper-parameter tuning. 
%We evaluate performance of these three Table QA models trained on WikiSQL dataset.  Thus, we can analyze the efficacy of open domain training based Table QA systems on our new dataset.
%To perform zero shot evaluation, we do not use any validation set from \sys, using instead the validation set from WikiSQL dataset for model selection.
% Our aim is to study the efficacy of open domain training based Table QA systems, when applied as it is on a specific domain like \sys. Therefore, \sys{} only has validation set and test set but no training set. So we use standard WIKISQL dataset as training set for all our experiments. 
We use the original source code released by the respective authors, pretrained weights along with details in their papers, for setting up all baseline models. In particular, for TaBERT \cite{tabert}, we use the pre-trained BERT released on the official GitHub repository\footnote{\url{https://github.com/facebookresearch/TaBERT}} with semantic parser MAPO\footnote{Source code available at \url{https://github.com/pcyin/pytorch_neural_symbolic_machines}}. TaBERT was trained for 10 epochs on 4 Nvidia Tesla v100s with batchsize of 10, number of explore samples as 10 and all other hyperparameters kept exactly the same as \cite{tabert}. For RCI~\cite{rci}, we use the code released with the paper\footnote{\url{https://github.com/IBM/row-column-intersection/}} to train the model for 2 epochs on 2 Tesla v100s, with learning rate 2.5e-5 and batch size $128$.
 For TaPaS \citep{tapas}, we use the model\footnote{\url{https://storage.googleapis.com/tapas_models/2020_08_05/tapas_wikisql_sqa_masklm_large_reset.zip}}  trained on the WikiSQL dataset from the official GitHub repository\footnote{\url{https://github.com/google-research/tapas}}. On WikiSQL dev set TaBERT gives an accuracy of $70.5\%$, TaPas $89.2\%$, and RCI $89.8\%$. 

\subsection{Transforming Table Structures}
\label{sec:transform}

Existing table QA models are based on open domain web tables. So they assume that the input tables contain flat column headers (i.e., a single row of column headers) and no row headers. Therefore, none of the existing baselines are built for handling complex column or row header hierarchies seen in \sys . So, we experiment with transformation operations on the table to maximize these baselines' performance on \sys. 
%\sys{} contains tables with complex structures such as row headers and header hierarchies. As existing Table QA models can not process such tables, we apply a couple of preprocessing tricks to convert these tables to simple form, consumable by these models. 

{\bf Base transformations} are first performed on \sys~ tables to render the tables compatible to the models as follows:
\begin{itemize}
    \item \emph{Row headers as Table cells in a new column:} Row headers are added as the first column of the table as regular body cells. We use a dummy text \textit{header} as the column header for this new column.
    \item \emph{Header hierarchies as flat headers}: Header hierarchies are flattened by concatenating parent header text with children text.
\end{itemize}

Note that these base transformations are designed to help the baseline models perform better than if we ran them on the raw table. % as each cell in the first column of the table explicitly contains the entire row header hierarchy (which was not present as such in the original table).
For instance, when converting the table of Figure~\ref{fig:example}, the cell on the left of the blue cell will contain the concatenated row header hierarchy (i.e., `Mainline passenger revenue miles (millions)'). This should help the models (which are not built to recognize row header hierarchies) perform better on \sys. 

{\bf Transposing tables.} However, after running the models, we observed that there was further room for improvement. In particular, 
we observed that in many tables in \sys, row headers contain more information than column headers. For example, in Figure~\ref{fig:example}, row headers contain the metric names, which are much more descriptive than the column headers containing just the year information. Based on this intuition, we experimented with transposing the headers, so that row headers become column headers (which the models are trained to pay more attention to) and vice versa (and body cells are appropriately transposed as well). This led to three versions of \sys{} data: (1) \emph{Base:} without transposing tables, (2) \emph{All transpose:} With all tables transposed, and (3) \emph{Partial transpose:} Transposing tables that have more characters in row headers than column headers. Table \ref{tab:transpose} depicts the accuracy of baseline models on each dataset version. Interestingly, RCI and TaBERT benefit from transposing tables, while the performance of TaPas declines. For our subsequent analysis, we pick for each model the version of the data that provides the highest performance for it.  

\setlength{\tabcolsep}{4pt}

\begin{table*}[t]
\centering
\begin{subfigure}{0.38\textwidth}
\begin{tabular}{l|ccc}
\toprule
\textbf{Version} & \textbf{TaBERT} & \textbf{TaPaS} & \textbf{RCI}\\
\midrule
Base & 33.20 & 49.32 & 40.58  \\
All T & 33.39 & 43.88 & 48.54 \\
Partial T & 33.98 & 46.80 & 51.84 \\
\bottomrule
\end{tabular}
\caption{Accuracy of Table QA on different transformations of the tables in \sys{} (\textbf{Base} = No transpose, \textbf{All T} = All transpose, \textbf{Partial T} = Partial Transpose).}
\label{tab:transpose}
\end{subfigure}
\begin{subfigure}{0.6\textwidth}
\centering
\begin{tabular}{l|ccc}
\toprule
\textbf{Data subset} & \textbf{TaBERT} & \textbf{TaPaS} & \textbf{RCI}\\
\midrule
 Overall accuracy & 33.98 & 49.32 & 51.84  \\
\midrule
KPI-driven & 41.37 & 48.26 & 60.00 \\
Table-driven & 31.08 & 50.0  & 48.64 \\
\midrule
Row header hierarchy & 21.92 & 47.26 & 45.89 \\
No row header hierarchy & 38.75 & 50.39 & 54.20 \\
\bottomrule
\end{tabular}
\caption{Accuracy of Table QA models on slices of \sys}
\label{tab:analysis}
\end{subfigure}
\caption{Accuracy of Table QA models on \sys}
\end{table*}

\subsection{Analyzing Baseline Performance on \sys{}'s Dimensions}
To gain further insights on how domain vocabulary, table structure, and question phrasing affect the performance of Table QA models, we next evaluate them on each of the three dimensions of our dataset: (1)~KPI-driven vs Table-driven, (2)~Row header hierarchy vs No Row header Hierarchy and (3)~Original vs Paraphrased questions. Table~\ref{tab:analysis} shows the results for the the first two, while Table~\ref{tab:analysis2} provides the results for the third dimension. We next discuss each dimension in detail:

%In this section we present the performance of each on the baselines on \sys. As mentioned in Section~\ref{sec:approach}, \sys{} has multiple dimensions of complexity including the unconventional table structure with row/column header hierarchies to specific domain vocabularies of airlines in the form of KPIs. Table~\ref{tab:analysis} compares the overall accuracy of each of the baselines on \sys{} and also breaks down the overall performance across three important axes of the dataset: (1)~KPI-driven vs Table-driven, (2)~Row header hierarchy vs No Row header Hierarchy and (3)~Original question vs Paraphrased. Below we discuss Table~\ref{tab:analysis} results in detail:

${\textbf{Overall Accuracy.}}$ Unlike existing datasets, such as WikiSQL and WikiTableQuestions with flat column headers and no row headers, \sys{} with its rich row and column header hierarchies, poses an additional challenge for state-of-the-art Table QA approaches. RCI framework is designed to give attention to individual rows and columns to extract the right intersection. Therefore, as expected, it performs best on \sys{} with 52\% accuracy. TaPas relies on an attention model as well, but considers the entire table as a whole, including its rows and columns. Therefore, TaPas performs comparable to RCI and with ~49\% accuracy. On the other hand, TaBERT takes a two step approach and is designed to produce intermediate logical forms to capture the complex intents from the question. Unlike RCI or TaPas, TaBERT does not have an end-to-end differentiable model for identifying right row(s) and column(s); therefore it performs worst on \sys{} with ~34\% accuracy.  

This relative performance trend, as witnessed in overall accuracy, holds also true when evaluating baselines on slices of the dataset along various dimensions. RCI and TaPas exhibit the best performance, while TaBERT trails further behind. 

\textbf{KPI-driven vs Table-driven.}
Table~\ref{tab:analysis} shows the performance of the baselines on KPI-driven vs Table-driven questions. As shown, accuracy is always higher for the former than the latter. The result can be correlated with the definitions of KPI-driven and Table-driven questions as mentioned in section~\ref{sec:approach} and indicates two key observations:

(1) There is a limited number of Airline KPIs, which most frequently exist as is in 10K documents submitted by an airline company. Therefore, even if a KPI name is an airline industry-specific term, the uniqueness of it helps each of the baselines to correctly identify the right answer.

(2) KPI-driven questions are formed by having a KPI in mind and searching for the corresponding term in the document. As a result, KPI-driven questions may be more natural than table-driven questions, which were formed by looking at a table and trying to form a question. As a result, it is much more common to find distorted utterances of row/column headers and/or cell values in table-driven questions, making it harder for the baseline to identify the correct answer from the question utterance. This observation has been also made in previous table-driven question annotation datasets, such as WikiTableQuestions. By containing both types of questions, \sys{} combines the more natural KPI-driven questions with the more challenging Table-driven questions. 

For the above reasons, KPI-driven questions tend to have better results than Table-driven questions. The gap is highest in RCI at 12\%, which is expected, as RCI can uniquely identify individual rows/columns using the KPI name.

\textbf{Row Header Hierarchy vs No Row Header Hierarchy.}
One of the key challenges associated with \sys{} are row/column header hierarchies. While we tried to help the baselines (which have not built with complex header structures in mind) deal with hierarchies through the table transformations described in Section \ref{sec:transform}, this implicit treatment of headers has two important limitations: 
%, which are quite common not only in most practical scenario of tables within reports and so on. However, none of the baselines are really trained to work with such tables. As described in our pre-processing step, although we embed such header hierarchies into row/column names before passing it to Table QA systems, it has two main limitations: 
(1) the explicit hierarchical information is lost and (2) in some cases, transformations may add noise into a row/column. Therefore, it is not surprising that questions that depend on row header hierarchies negatively affect the performance of all baselines and cause an average drop of ~13\%. \\
%RCI has the least drop of ~9\%, because of its ability to give attention to individual rows and columns. 

\begin{table*}[!t]
\centering
\begin{tabular}{l|ccc}
\toprule
\textbf{Paraphrase} & \textbf{TaBERT} & \textbf{TaPaS} & \textbf{RCI}\\
\midrule
All correct & 25.00 & 38.88 & 33.33  \\
Any correct & 29.17 & 30.55 & 34.72 \\
All wrong & 45.83 & 30.55 & 31.94 \\
\bottomrule
\end{tabular}
\caption{Percentage of paraphrased question sets that are (a) all correctly answered, (b) at least one correctly and another one incorrectly answered, and (c) all incorrectly answered by each baseline}
\label{tab:analysis2}
\end{table*}

${\textbf{Paraphrase Handling.}}$
Paraphrasing is an important aspect of \sys, because it allows us to see the effect of domain shift in the natural language understanding capability of Table QA systems. \sys{} contains $76$ paraphrased questions which can be grouped into $72$ paraphrased question sets (i.e., sets that include the original questions and all its paraphrases). Table~\ref{tab:analysis2} shows the effect of paraphrasing by breaking down the predictions of the baselines on the paraphrased question sets into three categories:
\begin{enumerate}[noitemsep]
    \item \textbf{All Correct:} When all questions in the set are answered correctly.
    \item \textbf{Any Correct:} When at least one question in the set is answered correctly and at least another question is answered incorrectly.
    \item \textbf{All Wrong:} When all questions in the set are answered incorrectly.
\end{enumerate}
The 2nd category (\textbf{Any correct}) is important to analyze from the point of view of a Table QA system's NLU capability. It essentially means that there exists a way of asking a question that leads to the right answer, implying that the Table QA system can handle such questions. Whereas, phrasing that same question in a different way leads to wrong results. This could be attributed to the domain shift. Since baselines are trained on the open-domain WikiSQL dataset, when tested on the domain-specific \sys, they can only handle certain ways of phrasing the questions, with almost ~30\% of the questions being supported when phrased in one way but not in a different way. This points to an important observation for practical domain shift scenarios: Table QA systems become sensitive to question phrasing in a significant way (in almost 30\% of cases).

%% file: 5_conclusion.tex
\section{Conclusion}
\label{sec:conclusion}
Table QA systems form a special type of QA systems that leverage information within tables to provide answers to natural language questions. Recent transformer-based Table QA approaches mainly innovate at the decoder stage to ensure that the tabular format is understood and leveraged. As a result, they provide high performance on existing Wikipedia-based datasets with simple tables. In this work, we present the first Table QA dataset that explicitly captures tables with complex structure, including column and row header hierarchies. Note, that while our dataset focuses on financial documents of the airline industry, such tabular structures are common in many other scientific and business documents. Our benchmarking of state-of-the-art Table QA methods show the deficiency of end-to-end, weakly-supervised and row-column encoding methods. We hope to encourage the community to consider new Table QA approaches that can support such complexity, so that Table QA methods can more effectively support a wider range of scientific and business use cases.